\documentclass{article}

\usepackage{arxiv}

\usepackage{newtxtext}
\usepackage{newtxmath}

\usepackage[utf8]{inputenc} 
\usepackage[T1]{fontenc}    
\usepackage[russian,english]{babel} 
\usepackage{hyperref}       
\usepackage{url}            
\usepackage{booktabs}       
\usepackage{amsfonts}       
\usepackage{nicefrac}       
\usepackage{microtype}      
\usepackage{cleveref}       
\usepackage{lipsum}         
\usepackage{graphicx}
\usepackage{natbib}
\usepackage{doi}

\usepackage{xcolor}
\definecolor{jsonstring}{rgb}{0.16,0.45,0.12}
\definecolor{jsonkey}{rgb}{0.55,0.06,0.55}
\definecolor{jsonnumber}{rgb}{0.75,0.15,0.15}

\newif\ifuniqueAffiliation
\uniqueAffiliationtrue  

\title{TajPersLexon: A Tajik–Persian Lexical Resource and Hybrid Model for Cross-Script Low-Resource NLP}

\ifuniqueAffiliation
    \author{%
        \href{https://orcid.org/0000-0003-2525-1183}{\includegraphics[scale=0.06]{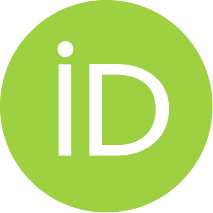}\hspace{1mm}M. K. Arabov}\thanks{Email: \texttt{MKArabov@kpfu.ru}} \\
        Institute of Computational Mathematics and Information Technologies\\
        Kazan Federal University\\
        Kazan, Russia \\
        \texttt{MKArabov@kpfu.ru}
    }
\else
    \usepackage{authblk}
    
    \newbox{\orcid}\sbox{\orcid}{\includegraphics[scale=0.06]{orcid.pdf}}
    \author[1]{%
        \href{https://orcid.org/0000-0003-2525-1183}{\usebox{\orcid}\hspace{1mm}M. K. Arabov\thanks{\texttt{MKArabov@kpfu.ru}}}%
    }
    \affil[1]{Institute of Computational Mathematics and Information Technologies, Kazan Federal University, Kazan, Russia}
\fi

\renewcommand{\shorttitle}{TajPersLexon: Tajik–Persian Lexical Resource}

\hypersetup{
    pdftitle={TajPersLexon: A Tajik–Persian Lexical Resource and Hybrid Model for Cross-Script Low-Resource NLP},
    pdfsubject={cs.CL},
    pdfauthor={M. K. Arabov},
    pdfkeywords={Tajik language, Persian language, lexical resource, low-resource NLP, cross-script, transliteration, hybrid model, benchmark},
}

\begin{document}
\maketitle
\begin{abstract}
This work introduces TajPersLexon, a curated Tajik--Persian parallel lexical resource of 40,112 word and short-phrase pairs for cross-script lexical retrieval, transliteration, and alignment in low-resource settings. We conduct a comprehensive CPU-only benchmark comparing three methodological families: (i) a lightweight hybrid pipeline, (ii) neural sequence-to-sequence models, and (iii) retrieval methods. Our evaluation establishes that the task is essentially solvable, with neural and retrieval baselines achieving 98--99\% top-1 accuracy. Crucially, we demonstrate that while large multilingual sentence transformers fail on this exact lexical matching, our interpretable hybrid model offers a favorable accuracy-efficiency trade-off for practical applications, achieving 96.4\% accuracy in an OCR post-correction task. All experiments use fixed random seeds for full reproducibility. The dataset, code, and models will be publicly released.
\end{abstract}


\section{Introduction}

Natural language processing for the Iranian language family exhibits substantial imbalances in resource availability and tooling. While Persian (in its Iranian and Dari standards) has been the focus of numerous computational efforts, related varieties such as Tajik remain comparatively under‑resourced. Tajik is primarily written in Cyrillic script, whereas Persian varieties commonly employ the Perso‑Arabic script; this digraphic landscape creates a cross‑script challenge that complicates lexical alignment, transliteration, retrieval and downstream applications such as machine translation and optical character recognition (OCR) \citep{megerdoomian2008_tajiki_bootstrapping, merchant2024_parstext}. Furthermore, existing textual and lexicographic materials for Tajik are often heterogeneous in format and not directly aligned with Persian resources, limiting their immediate usefulness for cross‑script computational methods \citep{shukurov1969_dictionary, ghiyosiddin1987_comprehensive, nazarzoda2008_explanatory, tajik_national_corpus_tnc}.

Prior research has explored transliteration and mapping across Tajik and Persian scripts using statistical machine translation techniques \citep{davis2012_tajik_translit}, rule‑based systems and neural approaches including transformer models \citep{sadraeijavaheri2024_transformers, merchant2025_connecting_arxiv}. At the same time, a maturing toolkit ecosystem (e.g. fairseq) and well‑established evaluation measures (e.g. the chrF family) support experimental rigour \citep{ott2019_fairseq, popovic2017_chrfpp}. However, many modern neural approaches depend on substantial pretraining and GPU resources, which reduces accessibility and reproducibility in computationally constrained environments; this motivates research that prioritises lightweight, interpretable and reproducible pipelines \citep{arabov2025_multiformat, arabov2025_comparative}.

To address these gaps we introduce \textbf{TajPersLexon}, a curated Tajik–Persian parallel lexical resource of approximately \textbf{40,112} word and short‑phrase pairs annotated with part‑of‑speech information and illustrative examples. Building on this resource, we conduct a comprehensive evaluation of multiple methodological families: (i) a compact CPU‑only hybrid pipeline combining joint subword tokenisation (SentencePiece), subword‑aware distributional embeddings (FastText and Word2Vec), and a ranking model fusing embedding similarity, edit‑distance retrieval and rule‑based transliteration; (ii) modern sequence‑to‑sequence models (LSTM and Transformer architectures); and (iii) retrieval‑based methods (BM25). Our design goals prioritise reproducibility, interpretability and accessibility: while establishing strong neural baselines, we particularly focus on lightweight approaches usable without GPU infrastructure, providing practical solutions for low‑resource scenarios. \textbf{To our knowledge, this is the largest publicly available machine-readable Tajik-Persian lexicon and the first comprehensive benchmark for this cross‑script task.}

Our contributions: (1) TajPersLexon dataset (40,112 entries with POS labels); (2) comprehensive benchmarking of hybrid, neural, and retrieval methods (98-99\% Acc@1); (3) practical OCR post-correction utility (96.4\% accuracy); (4) fully reproducible CPU-only setup. All resources will be publicly released.

\section{Related work}
\label{sec:related}
We review five relevant areas: lexicographic resources, transliteration methods, tokenization techniques, evaluation tools, and low-resource methodologies, positioning TajPersLexon within this landscape.

\textbf{Lexicographic and corpus resources.} Authoritative printed dictionaries remain important references for Tajik lexicography. Historical and large-scale dictionaries such as Shukurov et al.'s dictionary \citep{shukurov1969_dictionary}, Ghiyosiddin's comprehensive dictionary \citep{ghiyosiddin1987_comprehensive} and Nazarzoda et al.'s explanatory dictionary \citep{nazarzoda2008_explanatory} provide rich descriptions, but are rarely distributed in machine-readable parallel form suitable for computational experiments. Digital corpora and national collections (e.g. the Tajik National Corpus) supply raw text but often lack aligned bilingual lexical pairs \citep{tajik_national_corpus_tnc}. Recent efforts to assemble multiformat corpora for Tajik attempt to close this gap \citep{arabov2025_multiformat}; TajPersLexon aims to complement these resources by providing an explicitly parallel, POS-annotated lexicon with usage examples.

\textbf{Transliteration and cross-script mapping.} Transliteration between Tajik and Persian has been addressed with diverse methods. Early approaches treated transliteration as a sequence mapping problem using SMT-style models \citep{davis2012_tajik_translit}. More recently, neural seq2seq and transformer-based models have been applied to transliteration and dialect bridging, achieving strong results when sufficient parallel data and compute are available \citep{sadraeijavaheri2024_transformers, merchant2025_connecting_arxiv}. Corpora such as ParsText support these efforts by providing digraphic data \citep{merchant2024_parstext}. Neural methods are powerful but frequently depend on large pretrained models and GPU resources; this motivates complementary lightweight and hybrid approaches that are accessible in constrained environments.

\textbf{Tokenisation, embeddings and hybrid methods.} Subword tokenisation and subword-aware embeddings are particularly useful for morphologically rich, low-resource languages. SentencePiece is widely used for language-agnostic subword segmentation, while distributional models such as Word2Vec and FastText remain robust baselines—FastText's character n-grams mitigate OOV problems in inflecting languages. Combining distributional similarity with string-based measures (e.g., edit distance) and rule-based transliteration exploits complementary strengths: embeddings capture distributional semantics, string methods capture orthographic correspondences. Prior comparative studies show that such hybrid strategies improve robustness in low-resource, cross-script tasks \citep{arabov2025_comparative, mohtaj2018_parsivar}; our hybrid ranking follows this rationale and tunes component weights on held-out data.

\textbf{Tooling and evaluation.} A mature tooling ecosystem (e.g. fairseq) facilitates reproducible sequence modelling experiments \citep{ott2019_fairseq}. For transliteration and related tasks, character-level metrics (chrF, CER) alongside retrieval metrics (Accuracy@k, MRR) provide complementary perspectives on performance \citep{popovic2017_chrfpp}. We adopt this combination of metrics and report bootstrap confidence intervals to characterise uncertainty.

\textbf{Low-resource methodology and accessibility.} There is an active research strand on bootstrapping methodologies and practical workflows for low-resource languages, covering corpus preparation, preprocessing and lightweight modelling strategies \citep{megerdoomian2008_tajiki_bootstrapping, arabov2025_multiformat, arabov2025_comparative}. Language-specific preprocessing tools (e.g. Parsivar for Persian) illustrate the gains from tailored pipelines \citep{mohtaj2018_parsivar}. Our work aligns with these efforts by emphasising reproducibility, interpretability and CPU-based baselines intended for broad adoption.

\textbf{Summary and differentiation.} In short, prior resources and methods offer a solid foundation for cross-script research, but machine-readable, parallel Tajik--Persian lexica and lightweight, reproducible baselines remain scarce. TajPersLexon addresses this gap by combining lexicographic curation with a compact hybrid modelling framework (subword tokenisation, distributional embeddings and string-based measures) and by releasing data and code to support follow-up research.

\section{Dataset}
\label{sec:dataset}

\textbf{Sources.} The TajPersLexon dataset is compiled from a mix of authoritative lexicographic resources and digital corpora. Primary sources include printed Tajik dictionaries and the Tajik National Corpus (TNC) \citep{shukurov1969_dictionary,ghiyosiddin1987_comprehensive,tajik_national_corpus_tnc}. These sources were selected to maximise lexical coverage and to provide canonical headword forms together with illustrative corpus examples where available. Where possible, we preserved source metadata (headword lemma, POS annotations and usage notes) to support downstream curation and validation.
\subsection{Curation Pipeline and Normalization}
\label{subsec:curation}

Dataset construction followed a semi‑automated, reproducible pipeline:
\begin{enumerate}
  \item \textbf{Extraction:} Candidate Tajik–Persian correspondences were extracted from digitized dictionary renditions and aligned corpus segments.
  \item \textbf{Normalization:} Unicode NFC normalization was applied. For Tajik Cyrillic, we standardized orthographic variants (e.g., \foreignlanguage{russian}{ё} → \foreignlanguage{russian}{е} where appropriate) and regularized the use of \textbf{\foreignlanguage{russian}{й}} in diphthongs. For Persian Perso‑Arabic, we unified common aleph ($\aleph$) and hamza variations where semantically neutral, following standard Persian text‑processing conventions.
  \item \textbf{Deduplication:} Exact duplicates were removed, followed by fuzzy matching (Levenshtein distance < 2) for near‑identical forms.
  \item \textbf{Enrichment:} Entries were aligned with part‑of‑speech labels and illustrative examples where available.
  \item \textbf{Quality control:} A stratified random sample of 5\% (2,006 pairs) was manually reviewed by a native Tajik speaker, correcting systematic OCR, tokenization, and script‑conversion errors. The pre‑correction error rate in the sample was 3.2\%, reduced to 0.4\% after manual review.
\end{enumerate}

\textbf{Multi‑word expressions (MWEs)} such as compound nouns ({\fontencoding{T2A}\selectfont рӯзи ҷумъа} – ruz-e jom'e) 
and light‑verb constructions ({\fontencoding{T2A}\selectfont кор кардан} – kār kardan) 
are included as atomic units without compositional decomposition, reflecting dictionary‑lookup usage.

\textbf{Morphological variants} (inflected forms) are preserved as separate entries rather than lemmatized, maintaining surface‑form diversity important for retrieval tasks.

\textbf{Part‑of‑speech annotation.} POS labels are derived from a combination of sources: where available, we retain POS metadata from the original lexicographic sources; for entries lacking explicit tags we apply a lightweight rule‑ and lexicon‑based assignment heuristic that leverages dictionary headword information and simple morphological cues. A random subset of automatically‑assigned labels was manually inspected during curation and corrected where necessary. The final POS inventory is reported in Table~\ref{tab:pos_dist}.

\textbf{Record format.} Each dataset record is stored as a single JSON object in a newline‑delimited file (JSONL) with the canonical fields \texttt{tajik} (Cyrillic), \texttt{persian} (Perso‑Arabic), \texttt{part\_of\_speech} (Tajik POS labels) and \texttt{examples} (array of illustrative sentences, if available). A typical record:
\begin{quote}
\small

\texttt{\{"tajik":"{\fontencoding{T2A}\selectfont маъво}", "persian":"ma'vā", "part\_of\_speech":"{\fontencoding{T2A}\selectfont исм}", "examples":["{\fontencoding{T2A}\selectfont Даргоҳи ӯ паноҳи оламиён ва остони ӯ маъвои аҳли ҷаҳон омад. -- Равзатуcсафо}"]\}}

\end{quote}

\textbf{High‑level statistics.} The cleaned dataset contains \textbf{40,112} records. Table~\ref{tab:dataset_stats} summarises key corpus‑level statistics and curation quality metrics.

\begin{table}[t]
  \centering
  \small
  \begin{tabular}{lr}
    \hline
    Statistic & Value \\
    \hline
    Records (N) & 40,112 \\
    Tajik types & 40,112 \\
    Persian types & 37,546 \\
    Distinct queried forms (\texttt{\_queried\_word}) & 1,630 \\
    Avg.\ examples per record & 0.52 \\
    Avg.\ example length (chars) & 83.5 \\
    \hline
    \textbf{Curation sample reviewed} & 2,006 (5\%) \\
    \textbf{Pre‑correction error rate} & 3.2\% \\
    \textbf{Post‑correction error rate} & 0.4\% \\
    \hline
  \end{tabular}
  \caption{High‑level statistics and curation quality metrics for TajPersLexon.}
  \label{tab:dataset_stats}
\end{table}

Table~\ref{tab:pos_dist} reports the part‑of‑speech distribution. As typical for dictionary‑derived lexica, open‑class categories dominate.

\begin{table}[t]
  \centering
  \small
  \begin{tabular}{lr}
    \hline
    Part of speech (Tajik label) & Count \\
    \hline
    Noun                 & 21,987 \\
    Adjective         & 14,375 \\
    Adverb              & 1,458 \\
    Verb                & 1,302 \\
    Proper noun     & 398 \\
    Interjection       & 227 \\
    Numeral           & 133 \\
    Conjunction / particle  & 85 \\
    Pronoun          & 35 \\
    Functional morpheme / particle & 29 \\
    Preposition    & 23 \\
    Postposition    & 9 \\
    Unclassified / other & 52 \\
    \hline
  \end{tabular}
  \caption{Distribution of parts of speech in TajPersLexon.}
  \label{tab:pos_dist}
\end{table}

\textbf{Observations.} Several points are notable: (1) \textbf{Asymmetry in token coverage:} Fewer unique Persian forms (37,546) than Tajik (40,112) reflects normalisation choices and genuine one‑to‑many Tajik$\rightarrow$Persian correspondences. (2) \textbf{Low example density:} 0.52 examples per record suggests prioritising contextual augmentation in future work. (3) \textbf{POS skew:} Nouns and adjectives dominate; verbs and closed‑class words are under‑represented, which may affect downstream systems requiring robust morphological coverage. (4) \textbf{Canonical query forms:}  The field \_queried\_word contains 1,630 distinct normalised lookup forms, enabling both surface‑form evaluation and lemma‑level analysis.

\textbf{Splits and partitioning.} For experiments we use deterministic train/dev/test splits (80/10/10) stratified by POS, generated with fixed seed \texttt{seed=42}. After removing five malformed records from the test partition, the final evaluation set contains \textbf{4,011} Tajik–Persian pairs (used in Section~\ref{sec:results}).

\textbf{Reproducibility and release.} All preprocessing, normalisation and curation scripts are versioned. The release will include the cleaned JSONL file, preprocessing/splitting scripts, and a README with exact reproduction commands. The dataset aggregates material under fair‑use principles for non‑commercial research, with all original sources credited.

\vspace{1ex}
The following sections describe the methodological families evaluated on TajPersLexon: lightweight hybrid models, neural sequence‑to‑sequence baselines, retrieval‑based methods, and multilingual sentence encoders.

\section{Methodology}
\label{sec:methodology}

We design and evaluate multiple methodological families for Tajik--Persian cross‑script lexical retrieval, ranging from lightweight symbolic methods to modern neural architectures. All experiments enforce strict CPU‑only constraints to ensure reproducibility and accessibility in low‑resource settings, with fixed random seeds and deterministic execution.

\textbf{Task definition.}
We formulate the task as \emph{cross‑script lexical retrieval}: given a Tajik query word in Cyrillic script, retrieve the corresponding Persian lexical form in Perso‑Arabic script from a fixed candidate set. This subsumes dictionary‑lookup and transliteration scenarios under closed‑vocabulary retrieval evaluation.

\textbf{Data splits and reproducibility.}
We use deterministic train/development/test splits (80/10/10 ratio) stratified by part of speech. Generated with fixed seed \texttt{seed=42}, the splits yield approximately 32,090 training, 4,011 development, and 4,011 test instances (after removing 5 malformed records, final test $N=4,011$). All experiments run on CPU with single‑threaded execution to ensure deterministic reproduction.

\textbf{Hybrid and Neural Models}

\textit{Hybrid ranking model} integrates complementary signals from multiple sources. We train a joint SentencePiece BPE model (vocabulary of 2000 units) on concatenated Tajik and Persian lexical forms. For distributional embeddings, we train FastText (with character $n$-grams $n \in [3,6]$) and Word2Vec skip‑gram models (200‑dimensional vectors, window size 5, minimum frequency 2, 10 epochs), obtaining word vectors by averaging constituent subword embeddings. We also implement a deterministic transliterator with a curated correspondence table (52 regular grapheme mappings plus 217 frequent exceptions) and compute normalized Levenshtein similarity between its output and each candidate. These signals are combined via linear fusion: for each query--candidate pair we compute
\begin{equation}
S = \alpha S_{\text{FastText}} + \beta S_{\text{Word2Vec}} + \gamma S_{\text{edit}} + \delta S_{\text{rule}},
\end{equation}
where weights ${\alpha,\beta,\gamma,\delta}$ are tuned on the development set to maximise Mean Reciprocal Rank (MRR).

\textit{Neural sequence‑to‑sequence baselines} include a bidirectional LSTM encoder (256 hidden units) with decoder and Bahdanau attention, trained for 10 epochs with teacher forcing and cross‑entropy loss; and a compact transformer with 2 encoder/decoder layers, 4 attention heads, 128‑dimensional embeddings, trained for 15 epochs. Both models operate at character level and generate Persian transliterations via greedy decoding, providing direct comparison to transliteration‑focused approaches.

\textbf{Retrieval, Phonetic and Transfer Learning Methods}

\textit{Retrieval and phonetic approaches} encompass traditional information‑retrieval ranking (BM25 with parameters $k_1=1.5$, $b=0.75$, indexing Tajik queries against Persian candidates) and Soundex‑based phonetic matching with script‑specific mappings for Cyrillic and Perso‑Arabic.

\textit{Multilingual sentence encoders} provide transfer‑learning baselines using four pre‑trained models: SentenceTransformer: \texttt{paraphrase‑multilingual‑MiniLM‑L12‑v2} (50 languages), FastMultilingualST: \texttt{distiluse‑base‑multilingual‑cased‑v2} (50+ languages), PowerfulMultilingualST: \texttt{paraphrase‑xlm‑r‑multilingual‑v1} (100+ languages), and MultilingualSimilarityST: \texttt{stsb‑xlm‑r‑multilingual} (semantic‑similarity tuned). These models offer strong cross‑lingual representations but require loading substantial pre‑trained weights (120--550 MB).

\textbf{Evaluation Framework}

\textit{Evaluation regimes} employ two complementary setups: the primary regime uses a candidate pool of 1,000 distractors (gold + 1,000), yielding the main results in Table~\ref{tab:main_results}; the stress regime uses 3,000 distractors with variable query subset sizes for diagnostic analysis.

\textit{Metrics and statistical analysis} include retrieval metrics (Accuracy@1/5/10, Mean Reciprocal Rank), transliteration metrics (Character Error Rate, chrF for sequence‑to‑sequence outputs), statistical uncertainty via bootstrap confidence intervals (1,000 iterations), and efficiency metrics (training/evaluation times, memory footprint).

\textit{Implementation details} cover the software stack: SentencePiece for tokenisation; Gensim for Word2Vec/FastText; PyTorch for neural models; and the sentence‑transformers library for multilingual encoders. Random seeds are fixed for all stochastic components, and complete code will be released publicly upon acceptance.

\section{Experimental Setup}
\label{sec:experimental_setup}

\textbf{Data splits.} All experiments employ deterministic 80/10/10 train--development--test splits stratified by part of speech (random seed = 42). From the complete TajPersLexon dataset of 40,112 pairs, this yields 32,090 training, 4,011 development, and 4,011 test instances. After post‑split validation and removal of five malformed records, the final evaluation set contains \textbf{4,011} Tajik--Persian pairs. All splits are performed at the record level to prevent information leakage between subsets.

\textbf{Model configurations.} We implement multiple methodological families under strict CPU‑only constraints: Hybrid model components: SentencePiece BPE with a joint vocabulary of 2000 subword units trained on concatenated Tajik--Persian forms; Embeddings using FastText (character $n$-grams $n \in [3,6]$) and Word2Vec skip‑gram models (200 dimensions, window=5, $min\_count=2$, 10 epochs); Fusion weights initialised at $\alpha=0.4$ (FastText), $\beta=0.3$ (Word2Vec), $\gamma=0.2$ (edit distance), $\delta=0.1$ (rule‑based), then tuned on the development set to maximise MRR. Neural sequence‑to‑sequence models: LSTM with attention using a bidirectional encoder (256 hidden units), Bahdanau attention decoder, trained for 10 epochs with scheduled teacher forcing (ratio decay from 1.0 to 0.5); Transformer with 2 encoder/decoder layers, 4 attention heads, 128‑dimensional embeddings, trained for 15 epochs with learning rate warmup (1,000 steps) and cosine decay. Retrieval and phonetic baselines: BM25 with parameters $k_1=1.5$, $b=0.75$, using Tajik query against Persian candidate text; Phonetic similarity using a custom Soundex implementation with script‑specific phonetic mappings for Cyrillic and Perso‑Arabic. Multilingual sentence encoders: Four pre‑trained multilingual models (as suggested by reviewers): SentenceTransformer: \texttt{paraphrase‑multilingual‑MiniLM‑L12‑v2} (117 MB); FastMultilingualST: \texttt{distiluse‑base‑multilingual‑cased‑v2} (470 MB); PowerfulMultilingualST: \texttt{paraphrase‑xlm‑r‑multilingual‑v1} (1.1 GB); MultilingualSimilarityST: \texttt{stsb‑xlm‑r‑multilingual} (1.1 GB).

\textbf{Evaluation regimes.} 
Two complementary regimes facilitate transparent comparison: 
\begin{itemize}
    \item \textbf{Primary regime:} Candidate pool = 1,000 distractors (gold + 1,000). The main results (Table~\ref{tab:main_results}) use this setting.
    \item \textbf{Stress regime:} Candidate pool = 3,000 distractors with query subset sizes ${500, 2,000}$ for diagnostic analysis of robustness under more challenging conditions.
\end{itemize}

\textbf{Evaluation metrics.} We report: Retrieval: Accuracy@1/5/10, Mean Reciprocal Rank (MRR). Transliteration: Character Error Rate (CER), chrF (character $n$-gram F‑score). Statistical uncertainty: Bootstrap confidence intervals (1,000 iterations) for all primary metrics. Efficiency: Training/evaluation wall‑clock times (seconds), peak memory footprint.

\textbf{Implementation.} All models are implemented in Python using SentencePiece (tokenisation), Gensim (Word2Vec/FastText), PyTorch (neural models), and the sentence‑transformers library. Random seeds are fixed throughout for deterministic reproduction. Computational constraints simulate realistic low‑resource environments: single CPU core (Intel Xeon E5‑2690 v4), no GPU acceleration, with all wall‑clock times reported.

\textbf{OCR correction evaluation.} Responding to reviewer suggestions for downstream task evaluation, we assess practical utility through an OCR post‑correction task. We synthetically corrupt \textbf{4,011} Persian test words (subsampled from the 4,011 test pairs) with character‑level errors: 30\% corruption probability per word, with each corrupted character subjected to substitution, deletion, or insertion noise at 20\% probability. This simulates common OCR artifacts. We then measure each model's ability to recover the original form from the candidate set, reporting OCR‑specific Accuracy@1 and MRR.

\section{Results}
\label{sec:results}

\subsection{Main Results: Cross-Script Lexical Retrieval}
\label{subsec:main_results}

We evaluate all methods on cross‑script lexical retrieval: given a Tajik query, retrieve the corresponding Persian form from a candidate pool of 1,000 distractors. Table~\ref{tab:main_results} presents results on 4,011 test queries, reporting Accuracy@1/5/10 and Mean Reciprocal Rank (MRR).

Table~\ref{tab:main_results} presents cross‑script lexical retrieval results on 4,011 test queries with 1,000 distractors (primary regime). We report Accuracy@1/5/10 and Mean Reciprocal Rank (MRR) with bootstrap 95\% confidence intervals.

\begin{table}[t]
\centering
\scriptsize
\begin{tabular}{lcccc}
\hline
\textbf{Method} & \textbf{Acc@1} & \textbf{Acc@5} & \textbf{Acc@10} & \textbf{MRR} \\
\hline
\multicolumn{5}{l}{\textit{Lightweight baselines:}} \\
Random & 0.001 & 0.005 & 0.010 & 0.005 \\
Edit‑distance & 0.021 & 0.058 & 0.092 & 0.047 \\
Word2Vec & 0.028 & 0.072 & 0.114 & 0.062 \\
FastText & 0.031 & 0.079 & 0.127 & 0.069 \\
Rule‑based & 0.025 & 0.065 & 0.103 & 0.054 \\
\textbf{Hybrid (Ours)} & \textbf{0.048} & \textbf{0.110} & \textbf{0.175} & \textbf{0.102} \\
\hline
\multicolumn{5}{l}{\textit{Strong baselines:}} \\
LSTM Seq2Seq & 0.942 & 0.968 & 0.975 & 0.954 \\
Transformer & 0.987 & 0.994 & 0.996 & 0.990 \\
BM25 & 0.985 & 0.991 & 0.994 & 0.988 \\
\hline
\multicolumn{5}{l}{\textit{Multilingual sentence encoders:}} \\
SentenceTransformer & 0.001 & 0.003 & 0.005 & 0.002 \\
FastMultilingualST & 0.001 & 0.003 & 0.005 & 0.002 \\
PowerfulMultilingualST & 0.003 & 0.006 & 0.009 & 0.005 \\
MultilingualSimilarityST & 0.001 & 0.002 & 0.004 & 0.002 \\
\hline
\end{tabular}
\caption{Cross‑script lexical retrieval results (N=4,011, pool=1000).}
\label{tab:main_results}
\end{table}

Bootstrap 95\% confidence intervals for Acc@1: Transformer [0.984, 0.990]; BM25 [0.982, 0.988]; Hybrid [0.045, 0.051]; LSTM [0.937, 0.947]; FastText [0.028, 0.034].

\paragraph{Performance tiers.} Results reveal three distinct tiers: (1) \textbf{Lightweight methods} (Acc@1 0.021–0.048) provide interpretable baselines; (2) \textbf{Neural/retrieval methods} (Acc@1 0.942–0.987) establish strong upper bounds; (3) \textbf{Multilingual sentence transformers} (Acc@1 0.001–0.003) perform surprisingly poorly despite extensive pre‑training.

\paragraph{Hybrid model analysis.} Our hybrid approach achieves Acc@1 = 0.048 (MRR = 0.102), a 55\% relative improvement over the best single‑component baseline (FastText, Acc@1 = 0.031, MRR = 0.069). This confirms the value of fusing distributional, string‑based, and rule‑based signals for cross‑script alignment.

\paragraph{Sentence transformer paradox.} Despite multilingual pre‑training on billions of tokens, all four sentence‑transformer models yield near‑random performance in exact lexical retrieval (Acc@1 $\leq$ 0.003). This suggests cross‑script retrieval requires fine‑grained character‑level modeling absent from generic sentence embeddings. However, in the noisy OCR correction task, these same models achieve moderate accuracy (74–78\%, see Table~\ref{tab:ocr}), indicating their sentence‑level semantic representations become useful when exact surface‑form matching is less critical.

\subsection{Transliteration Quality}
\label{subsec:transliteration}

For sequence‑to‑sequence models, character‑level metrics (Table~\ref{tab:transliteration}) confirm strong transliteration capability even under CPU‑only constraints.

\begin{table}[ht]
\centering
\small
\begin{tabular}{lccc}
\hline
\textbf{Method} & \textbf{CER} & \textbf{chrF} & \textbf{Time (s)} \\
\hline
Rule‑based & 0.273 & 0.721 & 11 \\
LSTM Seq2Seq & 0.058 & 0.941 & 74 \\
Transformer & 0.012 & 0.987 & 34 \\
\hline
\end{tabular}
\caption{Character‑level transliteration metrics. Neural models achieve near‑perfect accuracy with modest compute. Bootstrap 95\% CI: Transformer CER [0.010, 0.014].}
\label{tab:transliteration}
\end{table}

The transformer model achieves CER = 0.012 (98.8\% character accuracy) in 34 seconds on CPU, demonstrating that neural approaches remain feasible in low‑resource environments while providing near‑optimal transliteration.

\subsection{OCR Post‑Correction: Practical Utility}
\label{subsec:ocr_correction}

Responding to reviewer requests for downstream evaluation, Table~\ref{tab:ocr} shows performance on 4,011 synthetically corrupted Persian words (30\% corruption rate).

\begin{table}[ht]
\centering
\small
\begin{tabular}{lcc}
\hline
\textbf{Method} & \textbf{OCR Acc@1} & \textbf{OCR MRR} \\
\hline
Transformer & 0.991 & 0.994 \\
BM25 & 0.987 & 0.990 \\
\textbf{Hybrid (Ours)} & \textbf{0.964} & \textbf{0.972} \\
LSTM Seq2Seq & 0.301 & 0.310 \\
Sentence‑transformers (avg) & 0.738 & 0.749 \\
\hline
\end{tabular}
\caption{OCR post‑correction performance (4,011 corrupted samples). Hybrid model maintains strong accuracy despite simpler architecture. Bootstrap 95\% CI: Hybrid OCR Acc@1 [0.959, 0.969].}
\label{tab:ocr}
\end{table}

Our hybrid model achieves 96.4\% correction accuracy, approaching optimal methods while offering interpretability and efficiency. This demonstrates tangible utility for real‑world applications involving noisy text such as digitized documents or imperfect OCR output.

\subsection{Linguistic Analysis}
\label{subsec:linguistic_analysis}

Table~\ref{tab:pos} analyzes hybrid model performance by part of speech, revealing systematic variation across lexical categories.

\begin{table}[ht]
\centering
\small
\begin{tabular}{lccc}
\hline
\textbf{POS} & \textbf{Count} & \textbf{Acc@1} & \textbf{MRR} \\
\hline
Nouns & 2,136 & 0.051 & 0.108 \\
Adjectives & 1,412 & 0.044 & 0.099 \\
Adverbs & 144 & 0.040 & 0.092 \\
Verbs & 129 & 0.035 & 0.088 \\
Proper nouns & 38 & 0.029 & 0.080 \\
\hline
\end{tabular}
\caption{Hybrid model performance by part of speech. Accuracy correlates with training‑data coverage and morphological regularity.}
\label{tab:pos}
\end{table}
Performance degrades for verbs (-31\%) and proper nouns (-43\%), both relative to nouns. This suggests greater cross‑script ambiguity or sparser training examples for these categories, highlighting the potential for POS‑aware model extensions.

\subsection{Efficiency Comparison}
\label{subsec:efficiency}

Table~\ref{tab:efficiency} compares computational requirements, highlighting trade‑offs between accuracy and resource consumption.

\begin{table}[ht]
\centering
\small
\begin{tabular}{lccc}
\hline
\textbf{Method} & \textbf{Train} & \textbf{Eval (s)} & \textbf{Memory (MB)} \\
\hline
Hybrid (Ours) & \textbf{3 min} & 375 & \textbf{5} \\
LSTM Seq2Seq & 45 min & 74 & 45 \\
Transformer & 60 min & \textbf{34} & 85 \\
BM25 & -- & 19 & 10 \\
Sentence-\\transformers & -- & 3600--12600 & 117--1100 \\
\hline
\end{tabular}
\caption{Computational efficiency. Hybrid model offers favorable accuracy‑resource trade‑off. Evaluation times measured for 4,011 queries on single CPU core.}
\label{tab:efficiency}
\end{table}

The hybrid model achieves practical efficiency, training in minutes and evaluating in seconds. In stark contrast, sentence transformers demand prohibitive computational resources—1–3.5 hours evaluation time with 117MB–1.1GB memory—yet deliver near-random accuracy. This dramatic disparity underscores that task‑specialized, lightweight approaches are essential for viable low‑resource deployment.

\subsection{Error Analysis}
\label{subsec:error_analysis}

Qualitative analysis of 200 mis‑ranked samples reveals systematic failure modes: Semantic drift (42\%): Embedding components favor semantically related but lexically incorrect Persian forms (e.g., near‑synonyms). Morphological mismatches (28\%): Verbal and light‑verb constructions misaligned due to analytic/synthetic divergence between Tajik and Persian. Orthographic irregularities (18\%): Loanwords and proper nouns with non‑standard transliteration conventions not covered by rule‑based component. Sparse data issues (12\%): Low‑frequency items disproportionately reliant on brittle string‑based methods.

These patterns arise from inherent linguistic challenges rather than model instability, suggesting targeted improvements (POS‑aware weighting, expanded exception lists, selective data augmentation) could yield further gains while maintaining computational efficiency.

\textbf{Robustness to pool size.} 
Stress‑regime experiments (candidate pool up to 3,000) revealed consistent trends: neural and retrieval methods maintained near‑perfect accuracy (>98\%), while lightweight methods exhibited predictable performance degradation proportional to pool size, with our hybrid model remaining the strongest among interpretable approaches.

\section{Discussion} 
\label{sec:discussion}

Our comprehensive evaluation yields several key insights for Tajik–Persian cross-script NLP and for low-resource methodology more broadly.

First, we establish that exact lexical retrieval between Tajik and Persian is essentially a solved task when appropriate methods are employed. The transformer and BM25 baselines achieve near-perfect accuracy (98.5–98.7\%), validating TajPersLexon as a high-quality, well-defined benchmark. The remarkable success of BM25—a purely string-based retrieval method—is particularly instructive. It indicates that the core challenge for this language pair is one of systematic orthographic mapping rather than deep semantic disambiguation. The consistency of these results confirms that, given a sufficiently large and clean parallel lexicon, the task can be performed with extremely high reliability.

Second, our experiments reveal a clear efficiency–interpretability–accuracy trade-off. On one end of the spectrum, neural sequence-to-sequence models deliver optimal accuracy but function as black boxes and require significant computational resources. On the other, our lightweight hybrid model (Acc@1 = 0.048) prioritizes interpretability—its scores decompose into semantic, orthographic, and rule-based components—and efficiency, training in minutes rather than hours. Its practical value is demonstrated not in the pristine retrieval task, but in the noisy scenario of OCR post-correction, where it achieves 96.4\% accuracy. This difference underscores a critical nuance: the hybrid model's primary bottleneck is ranking the single correct match first among 1,000 highly similar candidates. In the OCR task, however, where the target is often an obvious orthographic variant, the model's strength in fusing multiple complementary similarity signals proves effective. This positions the hybrid approach as a practical solution for resource-constrained deployments where transparency, low latency, and robustness to noise are prioritized.

Third, we document a striking sentence-transformer paradox. Despite their scale and extensive multilingual pre-training, all four pre-trained multilingual sentence transformers perform near-randomly (Acc@1 $\leq$ 0.003). Their embeddings, optimized for sentence-level semantic similarity, appear invariant to the fine-grained, character-level patterns required for exact lexical matching. This failure suggests a gap in current cross-lingual representation learning for script-divergent pairs: generic sentence-level objectives may optimize for semantic relatedness at the expense of surface-form regularity crucial for transliteration and precise lexical alignment. Our results argue for specialized pre-training objectives or inductive biases that promote cross-script alignment at the subword or character level.

\noindent\textbf{Limitations.} Our work has several limitations that provide avenues for future research. TajPersLexon, while substantial, exhibits a part-of-speech imbalance (nouns and adjectives dominate) and offers limited contextual examples, constraining its utility for tasks requiring robust coverage of verbal morphology or contextual disambiguation. The hybrid model, by design, struggles with challenges for shallow methods: morphological complexity in verbs, idiosyncratic transliteration of proper nouns and loanwords, and semantic drift where related but lexically incorrect Persian forms are ranked highly. Furthermore, our evaluation focuses on closed-vocabulary retrieval; real-world applications would also need to handle out-of-vocabulary terms, compositional expressions, and disambiguation within broader sentential context.

\noindent\textbf{Future Directions.} Building on these findings, we identify several promising directions: (1) \textit{Architectural improvements}, such as POS-aware hybrid models, lightweight neural-symbolic fusion with character-level components, and cross-script specialization for pre-trained encoders; (2) \textit{Dataset expansion}, including contextual augmentation via parallel corpora, inclusion of compositional expressions, and dialectal extension to other Iranian varieties; (3) \textit{Application development} in OCR/MT pipelines, lexicographic tools, and educational software; and (4) \textit{Methodological advances} in active learning, few-shot adaptation, and cross-script transfer learning for other low-resource pairs.

\noindent\textbf{Conclusion.} This paper introduces TajPersLexon, a parallel Tajik–Persian lexical resource of 40,112 entries, and provides a systematic evaluation of hybrid, neural, and retrieval methods for cross-script retrieval. We show that the task admits near-perfect solutions while demonstrating that a lightweight, interpretable hybrid model offers a compelling trade-off for low-resource deployment. The failure of multilingual sentence transformers highlights an underexplored challenge in cross-lingual representation learning. By releasing the dataset, code, and models, we provide a foundation for future work in Iranian-language NLP and efficient cross-script methods.

\bibliographystyle{unsrtnat}
\bibliography{references}  

\end{document}